\documentclass[letterpaper, 10 pt, journal, twoside]{IEEEtran} 

\IEEEoverridecommandlockouts                              % This command is only needed if 
\usepackage[T1]{fontenc}

\usepackage[ruled,vlined]{algorithm2e}

\usepackage{graphicx}
\usepackage{textcomp}
\usepackage{xcolor}
\usepackage{hyperref}
\usepackage[position=bottom]{subfig}%the position=bottom ensures the captures are at the bottom of the figures

\usepackage{tkz-graph}
\usetikzlibrary{shapes.geometric}

\usepackage{tikz}
\usetikzlibrary{decorations.pathreplacing}

\usepackage{array}
\usepackage{tabularx}
\newcolumntype{b}{X}
\newcolumntype{s}{>{\hsize=.4\hsize}X}
\newcolumntype{m}{>{\hsize=.7\hsize}X}
\newcolumntype{t}{>{\hsize=.3\hsize}X}
\newcolumntype{v}{>{\hsize=.2\hsize}X}
\newcolumntype{q}{>{\hsize=.1\hsize}X}

\usepackage{xcolor}

\usepackage{makecell}%for a wider alternative to \hline

\usepackage{amsfonts}%for mathfrak

\usepackage{amsmath}
\usepackage{amssymb}%for triangleq etc

\usepackage{float}

\usepackage{threeparttable}
\usepackage{soul}

\makeatletter
\newcommand\notsotiny{\@setfontsize\notsotiny\@vipt\@viipt}
\makeatother

\title{Few-shot Quality-Diversity Optimization}

\author{Achkan Salehi$^{\dagger}$, Alexandre Coninx$^{\dagger}$, Stephane Doncieux$^{\dagger}$%
\thanks{Manuscript received September 9, 2021; accepted January 28, 2022. This letter was recommended for publication by Associate Editor T. Horii and Editor T. Ogata upon evaluation of the reviewers' comments.}
\thanks{This work was supported by the European Union's H2020-EU.1.2.2 Research and Innovation Program through FET Project VeriDream under Grant Agreement Number 951992.}
\thanks{$^{\dagger}$ Sorbonne Universit\'e, CNRS, ISIR, F-75005 Paris, France. Email: {\texttt\footnotesize\{achkan.salehi, alexandre.coninx, stephane.doncieux\}@sorbonne-universite.fr}}
\thanks{\copyright 2022 IEEE.  Personal use of this material is permitted.  Permission from IEEE must be obtained for all other uses, in any current or future media, including reprinting/republishing this material for advertising or promotional purposes, creating new collective works, for resale or redistribution to servers or lists, or reuse of any copyrighted component of this work in other works.}
\thanks{Final version reference and link: A. Salehi, A. Coninx and S. Doncieux, "Few-shot Quality-Diversity Optimization," in IEEE Robotics and Automation Letters, doi: 10.1109/LRA.2022.3148438. \url{https://ieeexplore.ieee.org/document/9705622}}
}

\begin{document}
\markboth{IEEE Robotics and Automation Letters. Preprint Version. Accepted January 2022}
{Salehi \MakeLowercase{\textit{et al.}}: Few-shot Quality-Diversity Optimization}
\maketitle
\begin{small}

%===============================================================================

  %In the past few years, a considerable amount of research has been dedicated to the exploitation of previous learning experiences through Few-shot and Meta Learning, in problem domains ranging from Computer Vision to Reinforcement Learning based control. A notable exception where, to the best of our knowledge, no efforts have been made in this direction is Quality-Diversity (QD) optimization, which remains costly due to the reliance of QD methods on inherently sample inefficient evolutionary processes. This is unfortunate, since QD methods have been shown to be effective tools in dealing with deceptive minima and sparse rewards in Reinforcement Learning. We show that, given examples from a task distribution, information about the paths taken by optimization in parameter space can be leveraged to build a prior population, which when used to initialize QD methods in unseen environments, allows for few-shot adaptation. Our proposed method does not require backpropagation. It is simple to implement and scale, and furthermore, it is agnostic to the underlying models that are being trained. Experiments carried in both sparse and dense reward settings using robotic manipulation and navigation benchmarks show that it considerably reduces the number of generations that are required for QD optimization in these environments.
\begin{abstract}
  In the past few years, a considerable amount of research has been dedicated to the exploitation of previous learning experiences and the design of Few-shot and Meta Learning approaches, in problem domains ranging from Computer Vision to Reinforcement Learning based control. A notable exception, where to the best of our knowledge, little to no effort has been made in this direction is Quality-Diversity (QD) optimization. QD methods have been shown to be effective tools in dealing with deceptive minima and sparse rewards in Reinforcement Learning. However, they remain costly due to their reliance on inherently sample inefficient evolutionary processes. We show that, given examples from a task distribution, information about the paths taken by optimization in parameter space can be leveraged to build a prior population, which when used to initialize QD methods in unseen environments, allows for few-shot adaptation. Our proposed method does not require backpropagation. It is simple to implement and scale, and furthermore, it is agnostic to the underlying models that are being trained. Experiments carried in both sparse and dense reward settings using robotic manipulation and navigation benchmarks show that it considerably reduces the number of generations that are required for QD optimization in these environments. Our code is available at \url{https://github.com/salehiac/FAERY-original}.
\end{abstract}

\begin{IEEEkeywords}
Evolutionary Robotics; Reinforcement Learning; Transfer Learning
\end{IEEEkeywords}

\section{Introduction}

% Drop letter for first word of the Introduction
\IEEEPARstart{M}{any} engineering problems can be formulated as optimization problems, where the goal is to find a set of optimal or near-optimal parameters for a model. Obtaining such solutions is often made difficult by the presence of deceptive optima in the loss/fitness\footnote{The term fitness, borrowed from the evolutionary computation literature, is synonymous with return. In this paper, it specifically refers to discounted cumulative reward.} landscape, and in the case of Reinforcement Learning by sparse rewards. In such situations, methods that solely rely on gradient or higher-order local signals tend to fail as a result of poor exploration.

  In part because of their natural ability to deal with those issues, Quality-Diversity \cite{cully2017quality} (QD) methods have garnered considerable interest in the past few years, and have been applied to many problems, ranging from robotics \cite{cully2015robots,kim2021exploration,bossens2020qed} to video games \cite{charity2020baba,gravina2019procedural} and open-ended learning \cite{wang2019paired, wang2020enhanced}. The aim of these methods is to obtain a diverse set of well-performing solutions to a given problem. To do so, they not only aim to optimize the fitnesses of the agents in the population, but also maximize their \textit{behavioral diversity} according to some (learned or hand-designed) behavior descriptor space. This results in divergent exploration of parameter space, which prevents the search from getting trapped in deceptive minima. Unfortunately, the underlying evolutionary processes on which QD approaches almost always rely make them highly data-inefficient, and they are thus expensive to train from scratch. Some works such as QD-RL \cite{cideron2020qd} address the sample-inefficiency problem in dense reward settings by replacing evolutionary algorithms with policy-gradient methods and optimizing the population jointly for behavioral novelty and expected return. However, this does not work with sparse rewards and the resulting policies do not readily transfer to novel environments and need to be trained from scratch for each new task. 
  
  In this work, we consider a complementary but orthogonal direction that can handle both sparse and dense rewards. Our aim is, given a training set sampled from a task distribution $\mathcal{T}$, to learn a prior population that when used to initialize a QD algorithm to solve an unseen task $t_{new} \sim \mathcal{T}$ will require only few generations of training (instead of the typical hundreds or thousands). Our proposed approach, which we call FAERY, for \textbf{F}ew-shot Qu\textbf{A}lity-Div\textbf{ER}sit\textbf{Y} Optimization, maintains a prior population $\mathcal{P}$ that is used to initialize independent QD optimizations, and that is continuously refined by maximizing two meta-objectives. The latter are computed based on statistics gathered from the paths taken by the evolutionary processes in each of the QD instances. As the resulting prior $\mathcal{P}$ is used to improve the efficiency of learning, FAERY falls under the broad definition of meta-learning as "learning to learn", although it would be excluded by stricter, more recent definitions \cite{hospedales2020meta}. Note that the proposed method does not require gradients, which makes it suitable for sparse reward environments. Furthermore, it has the appealing characteristic of being easy to implement and scale.

  We perform experiments on navigation tasks with sparse rewards in randomly generated mazes, and on robotic manipulation tasks from the Meta-World \cite{yu2020meta} benchmarks where we use the dense reward to accelerate the policy search. We distinguish intra-task generalization from cross-task generalization, both of which will be formally defined in section \ref{sec_experim}. While the primary focus of the conducted experiments is on few-shot adaptation in the latter case, we also present results indicating that using the priors learned by the proposed approach on a source task $t_{source}$ allows effective knowledge transfer to target tasks, especially in comparison to direct transfer of expert policies from $t_{source}$ to the target environments.

  We set the context and formalize our objectives in the following section (\S\ref{formulation_sec}). Section \ref{sec_related} is dedicated to the positioning of our work with respect to the literature. We describe FAERY in section \ref{sec_faery} and present experimental evaluations on navigation tasks and meta-world in \ref{sec_experim}. Closing discussions and remarks are the subject of \ref{sec_concl}.

\section{Problem Formulation and Notations}
  \label{formulation_sec}
  
  We place ourselves in a Reinforcement Learning context and assume that each environment $t_i$ sampled from a task distribution $\mathcal{T}$ can be modeled as a Partially Observable Markov Decision Process (POMDP) noted $<\mathcal{S}, \mathcal{A}, r, \rho, \gamma, \Omega, \rho_{obs}, \rho_{init}>$ where $r$ is a reward function, $\rho: \mathcal{S}\times \mathcal{A} \times \mathcal{S} \rightarrow [0,1]$ defines the transition probabilities $\rho(s_{t+1}, a_{t},s_t)=p(s_{t+1}| s_t, a_t)$ and $\rho_{obs}(o,s_t,a_t)=p(o|s_t,a_t)$ is the probability of and element of the observation space $\Omega$ given a state-action pair. The scalar $\gamma \in [0,1]$ is a discount factor and $\rho_{init}$ specifies the distribution over initial states. In this work, we consider that the initial state is fixed up to an infinitesimal perturbation, \textit{i.e.} $\rho_{init}$ is a Gaussian $\mathcal{N}(s_0,\epsilon)$. 

  As is usual in the literature \cite{cully2017quality, pugh2016quality, lehman2011evolving}, we broadly define the term Quality-Diversity (QD) as referring to any algorithm that combines fitness-based optimization with divergent exploration and produces a \textit{behaviorally diverse} set of solutions that are locally optimal according to some algorithm-dependent notion of neighborhood. For example, in hexapod robot locomotion, one could consider walking speed as the fitness to optimize and the gait as the behavior \cite{cully2015robots}. A QD algorithm would ideally construct a set of controllers that, given a desired gait (\textit{e.g.} due to damaged limbs) will be able to walk as fast as possible, resulting in increased robustness to environmental changes. It should be noted that in some cases as in navigation problems with sparse rewards \cite{lehman2011abandoning}, this diverse set of solutions \textemdash often referred to as an archive in the literature \textemdash is only useful to avoid convergence to deceptive local minima, and as a result is a byproduct of the algorithm, not the end goal. 
  
  To compute behavioral diversity, we assume the existence of a (hand-designed or learned) behavior function $\mathcal{B}(\tau) \mapsto \mathfrak{B}$ which maps each trajectory $\tau=\{(s_i, a_i)\}_i$ to a metric behavior space $\mathfrak{B}$. Based on that space, different criteria of behavioral diversity such as Novelty \cite{lehman2011abandoning}, Surprise \cite{gravina2016surprise} or Curiosity \cite{cully2017quality} can be considered. Note that Quality-Diversity methods usually enable dynamic exploration by storing a number of previously encountered behaviors and agents in an archive, which can be structured \cite{mouret2015illuminating} or structureless \cite{lehman2011evolving}. However, which of these two categories an algorithm falls into and which diversity metric it uses is irrelevant to our work. This is because we address a drawback that is common to all the aforementioned QD methods, namely, their reliance on sample-inefficient evolutionary processes.

 Given a problem $t \sim \mathcal{T}$, a typical QD instance will run for $g \in \{1, ..., \mathcal{G}_{QD}\}$ generations until the evaluation budget $\mathcal{G}_{QD}$ is exceeded, or a satisfaction condition (\textit{e.g.} coverage and average fitness) is reached at some generation $\mathcal{G}^{*}$. Note that $\mathcal{G}^{*}$ is a random variable. In what follows, we will note

  \begin{equation}
    \hat{\mathcal{G}}(t)=\mathbb{E}[\mathcal{G}^{*}]
  \end{equation}

  the expected number of generations in which a given QD method reaches the satisfaction conditions on an environment $t \sim \mathcal{T}$.
  
 For an unseen problem $t_{n}\sim \mathcal{T}$, the value $\hat{\mathcal{G}}(t_n)$ is in practice often very high. Assuming a training set $\mathcal{T}_{train}$ with examples from $\mathcal{T}$, we formulate our aim as learning a prior $\omega$ that if taken into account by the QD algorithm will significantly decrease the number of generations that are required to solve $t_n$:
 
  \begin{equation}
    \mathbb{E}[\mathcal{G}^{*}|\omega]\ll \hat{\mathcal{G}}(t_n).
    \label{eq_goal}
  \end{equation}

In this paper, we assume that a binary signal $\psi_i$ can be associated to each task $t_i$ such that 

  \begin{equation}
    \psi_{i}(\theta_l)=
    \begin{cases}
      1 & \text{if } \theta_l \text{ solves } t_i\\
      0 & \text{otherwise}
    \end{cases}
  \end{equation}holds.

  This is a small restriction, as most problems can be formulated in this way. The objective of tasks that do not have clear goal states/behaviors usually is either return maximization or pure exploration. In those cases, one can set $\psi_{i}(\theta_i)=1$ if the returns or the quality of exploration (according to some well-defined criterion) exceed some threshold.

Throughout all sections, $\Delta(\Theta, \lambda)$ will denote a mutation operator that will return $\lambda$ policies resulting from perturbations made to the parameters of the population $\Theta$, according to some mutation scheme. Furthermore, $\mu$ and $\lambda$ will respectively denote population and offspring size.
  
\section{Related Work}
\label{sec_related}

The proposed approach is naturally related to previous works on data-efficient QD as well as to the literature on few-shot and meta-learning.

 \textbf{Data-efficient QD.} In general, most of those works fall into one of two categories. The first one is composed of methods that attempt to improve the data-efficiency of the evolutionary processes used in QD algorithms by using information from the reward signal or from diversity criteria to bias the sampling of new individuals from a parent population. While traditional QD methods do select the population based on fitness and diversity, they apply random mutations to their weights in a manner that is independent from the problem at hand, which is nearly always suboptimal. Recently, \cite{fontaine2020covariance, paolo2021sparse} have proposed to combine populations of CMA-ES \cite{iruthayarajan2010covariance} instances (often called emitters in the literature) with QD algorithms, and both of these works show increased efficiency in sparse reward settings. Focusing on problem domains with dense rewards, QD-RL \cite{cideron2020qd}, replaces sample-inefficient evolutionary algorithms with policy-gradient methods and optimizes the population jointly for behavioral novelty and expected return. In a similar manner, ME-ES \cite{colas2020scaling} leverages the more efficient ES strategies to replace the random exploration of vanilla MAP-elites. In other recent works \cite{shi2020efficient}, in addition to selecting and mutating most novel individuals and in order to accelerate behavior space coverage, the authors propose to push less novel individuals toward the most novel ones by creating targets in behavior space, in a manner reminiscent of Goal Exploration Processes \cite{forestier2017intrinsically}. 
 
Methods from the second family do not directly address the low sample-efficiency of the evolutionary methods. Instead, they attempt to model and learn different components from the decision process of interest in a data-efficient manner. For example, learning a model of the transition probabilities can improve data-efficiency on a real-word system: first, such a model allows sample-inefficient exploration to be carried out in simulation, thus reducing the need for evaluations on the real system. Second, it can be used to guide the search to areas in which the model expects higher returns. An example is the SAIL algorithm \cite{gaier2018data}, where costly fitness evaluations are replaced by calls to a surrogate function that is estimated via Bayesian Optimization, which is both guided (to high reward regions) and exploited (for final predictions) by MAP-elites instances. In M-QD \cite{keller2020model}, the authors propose to learn a model that predicts the behavior and quality given an individual, thus avoiding parameter space areas that are associated with low quality or low behavioral novelty. Note that many of the problems that are addressed by this second family of approaches have also been studied in the context of model-based Reinforcement Learning \cite{moerland2020model}, from which many ideas can be directly applied to QD optimization.

We emphasize that our proposed method is orthogonal to the approaches discussed above, as we address the construction of rapidly adaptable priors based on previous experience. That is, FAERY can be combined with the aforementioned approaches.

  \textbf{Few-shot learning and meta-learning.} In supervised learning, many approaches to few-shot learning have been developed, ranging from using simple hand-crafted label signatures as priors \cite{romera2015embarrassingly} to more complex metric-learning \cite{dong2018few} or meta-learning based \cite{vinyals2016matching} methods. In Reinforcement Learning, few-shot learning is almost always approached via meta-learning \cite{finn2017model, wang2016learning, santoro2016meta, duan2016rl, song2019maml, xu2018meta, nichol2018first}. Earlier definitions of meta-learning include any algorithm that changes the meta-parameters of another learning algorithm, such that the performance of the latter is improved \cite{Schaul:2010}. Under these definitions as well as more recent but similarly broad ones \cite{jospin2020hands}, approaches such as hyperparameter tuning via cross-validation \cite{bergstra2012random} would fall into the meta-learning category. Thus, some recent attempts at providing a definition add some restrictions, such as the requirement for training conditions to replicate testing conditions and/or the requirement for end to end optimization of the inner algorithm with respect to the outer objective \cite{hospedales2020meta}. The method we propose aims at improving the efficacy and efficiency of learning by exploiting meta-data associated with distinct training processes, but decouples the optimization of the meta-objectives from the inner optimization. As such, we feel that it falls in between the aforementioned definitions. Whether we consider FAERY as meta-learning or not, it presents several advantages. First, in contrast with most works (\textit{e.g.} \cite{finn2017model, xu2018meta, wang2016learning, oh2020discovering, duan2016rl, santoro2016meta}), it does not require gradients. This not only makes the method more suitable for deceptive or sparse reward problems, but also leaves more freedom in the design and addition of additional meta-objectives. Second, like \cite{finn2017model, song2019maml}, it is agnostic to the underlying models that are being trained. Finally, it is simple to implement and scale.

\section{FAERY}
\label{sec_faery}

Given a task distribution $\mathcal{T}$, our aim is to leverage learning experiences on tasks that are sampled from it to learn a population of prior policies $\mathcal{P}=\{p_1, ..., p_{\mu}\}$. Those priors should be distributed in parameter space in a manner that facilitates future QD optimizations on new tasks $t_n \sim \mathcal{T}$. In other words, setting $\omega\triangleq \mathcal{P}$ should satisfy equation \ref{eq_goal}. We learn these priors by optimizing two meta-objectives, which intuitively quantify an agent's polyvalence and adaptation speed.

Our proposed method is summarized in algorithm \ref{algo_pseudo}. Each iteration of the outer loop optimizes the prior population based on statistics gathered from $M$ distinct QD instances, that we will note $\{Q_i\}_{i=0}^{M-1}$. Note that given $i\neq j$, two instances $Q_i$ and $Q_j$ will in general run on two different environments $t_i, t_j$ sampled from the same task distribution $\mathcal{T}$. Each $Q_i$ will be initialized with a population $\mathcal{P}\cup \Delta(\mathcal{P},\lambda)$, and will optimize that population during $G_{QD}$ generations, using an evolutionary algorithm. The paths taken by the evolutionary process during the execution of $Q_i$ can be represented as a forest, composed of $\mu$ trees, such that the root of the $i-th$ tree corresponds to the policy $p_i$ from the prior population. This allows each solution found by the QD algorithm to be uniquely mapped to a single parent in the prior population\footnote{In this formulation, we have only considered QD algorithms that solely use mutations for evolution, and do not rely on crossover operations. The reasons for this choice are twofold: first, it simplifies notations. Second, crossover is less frequently used in the current literature. Nonetheless, extending our approach to handle crossovers is straightforward: each solution in the tree will result in updates to multiple parents instead of updates to a single one. } . An example of such a forest is given in figure \ref{fig_tree}.

Noting $\xi_{ij}$ the set of descendants of $p_j$ that result from the evolution process of $Q_i$, we define the two following fitnesses for each policy $p_j \in \mathcal{P}$:

\begin{equation}
  \begin{cases}
    f_0(p_j)=\sum\limits_{i=0}^{M-1} \sum\limits_{s \in \xi_{ij}} \psi_i(s)\\
    f_1(p_j)=\frac{-1}{f_0(p_j)}\sum\limits_{i=0}^{M-1} \sum\limits_{s \in \xi_{ij}} \psi_i(s) d_m(s, p_j).
  \end{cases}
  \label{eq_updates}
\end{equation}

where $d_m(s, p_j)$ returns the expected number of mutations necessary to reach $s$ from $p_j$. We approximate that value with the depth of $s$ in the forest associated to $Q_i$. The first objective in that equation, $f_0$, is the cumulated number of solutions that have evolved from $p_j$ over all optimizations $\{Q_i\}_{i=0}^M$. Regarding the second objective, the value $-1\times f_1$ corresponds to the number of mutations that are necessary to reach a solution from $p_j$, averaged over all environments $\{Q_i\}_{i=0}^M$. Both $f_0$ and $f_1$ are maximized.

Once those objectives are computed, any Pareto Optimization scheme (\textit{e.g.} NSGA2 \cite{deb2002fast}) can be used to select the best performing individuals from $\mathcal{P}\cup \Delta(\mathcal{P},\lambda)$. Those individuals will replace the ones in $\mathcal{P}$, and thus constitute the new prior that will be used in the next iteration of the algorithm.

The motivation behind the definitions given for the two fitness functions is perhaps better explained via a comparison with the way in which optimization-based meta-learning algorithms such as MAML \cite{finn2017model} operate. In the latter, adaptability to many tasks is enabled by computing the meta-update based on the sum of the losses from different sampled tasks, and quick adaptation of the prior is encouraged by limiting the number of gradient updates and sample-splitting. In FAERY, those two roles are respectively carried out by the maximization of $f_0$ and $f_1$: maximizing $f_0$ drives the $p_i$ to areas of parameter space where they can serve as priors for an increased number of tasks, and maximizing $f_1$, which minimizes the number of mutations, acts in a manner that is similar to reducing the number of gradient updates. Contrary to MAML-like approaches, however, the proposed optimization scheme can be used in both sparse and dense reward settings, as it does not rely on gradients.

%From an intuitive point of view, and using terminology from the evolutionary computation literature, the aim of FAERY can be seen as the occupation of evolvable, task-correlated niches. In practice, it can often happen that policies that are successful in solving a given task form clusters in parameter space. For example, this can be observed on some environments from the metaworld dataset: figures \ref{fig_tree}(b, c) both show 2d projections of the parameters (obtained with t-sne \cite{van2008visualizing}) of policies that are successful on (respectively) the \texttt{basketball-v2} and \texttt{soccer-v2} environments from that benchmark. Similar visualizations result from projecting successful policies in navigation tasks. In problems where such clusters exist, the aim of FAERY can intuitively be interpreted as learning a set of policies $\{p_j\}_j$ whose distribution in parameter space closely matches the distribution of the clusters. In the ideal case, there would be at least one policy $p_j$ per cluster. Each of those $p_j$ would then easily, through a few mutations, transform into a subset of other policies from that cluster. Using $\mathcal{P}$ in future QD optimizations will thus ensure that exploration starts in areas of parameter space that are associated to high rewards and/or interesting behaviors.

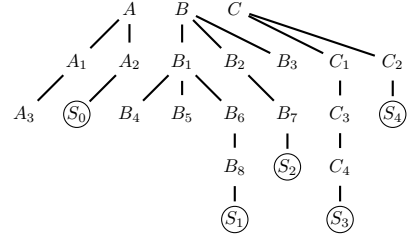
\begin{figure}[ht]
  \vspace*{0.1cm}
  \centering
  {\label{figtik}
  \begin{tikzpicture}[scale=0.7, transform shape]
    \GraphInit[vstyle=Empty]
    %======== Vertices
    %=== depth 0
    \Vertex[x=-1,y=0]{$A$}
    \Vertex[x=0,y=0]{$B$}
    \Vertex[x=1,y=0]{$C$}
    %=== depth 1
    %=
    \Vertex[x=-2,y=-1]{$A_1$}
    \Vertex[x=-1,y=-1]{$A_2$}
    %=
    \Vertex[x=0,y=-1]{$B_1$}
    \Vertex[x=1,y=-1]{$B_2$}
    \Vertex[x=2,y=-1]{$B_3$}
    %=
    \Vertex[x=3,y=-1]{$C_1$}
    \Vertex[x=4,y=-1]{$C_2$}
    %=== depth 2
    \Vertex[x=-3,y=-2]{$A_3$}
    \Vertex[x=-2,y=-2]{$S_0$}
    \Vertex[x=-1,y=-2]{$B_4$}
    \Vertex[x=0,y=-2]{$B_5$}
    \Vertex[x=1,y=-2]{$B_6$}
    \Vertex[x=2,y=-2]{$B_7$}
    \Vertex[x=3,y=-2]{$C_3$}
    \Vertex[x=4,y=-2]{$S_4$}
    %=== depth 3
    \Vertex[x=1,y=-3]{$B_8$}
    \Vertex[x=2,y=-3]{$S_2$}
    \Vertex[x=3,y=-3]{$C_4$}
    %=== depth 4
    \Vertex[x=3,y=-4]{$S_3$}
    \Vertex[x=1,y=-4]{$S_1$}

    %======== Edges 
    %=== depth 1
    %=
    \Edges($A_1$,$A$)
    \Edges($A_2$,$A$)
    %=
    \Edges($B_1$,$B$)
    \Edges($B_2$,$B$)
    \Edges($B_3$,$B$)
    %=
    \Edges($C_1$,$C$)
    \Edges($C_2$,$C$)
    %=== depth 2
    \Edges($S_0$,$A_2$)
    \Edges($A_3$,$A_1$)
    \Edges($B_4$,$B_1$)
    \Edges($B_5$,$B_1$)
    \Edges($B_6$,$B_1$)
    \Edges($B_7$,$B_2$)
    \Edges($S_4$,$C_2$)
    %=== depth 3
    \Edges($S_2$,$B_7$)
    %=== depth 4
    \Edges($C_3$,$C_1$)
    \Edges($C_4$,$C_3$)
    \Edges($S_3$,$C_4$)
    \Edges($B_8$,$B_6$)
    \Edges($S_1$,$B_8$)
    \node at (1,-4) [draw,circle,minimum height=0.5cm] {};
    \node at (3,-4) [draw,circle,minimum height=0.5cm] {};
    \node at (-2,-2) [draw,circle,minimum height=0.5cm] {};
    \node at (2,-3) [draw,circle,minimum height=0.5cm] {};
    \node at (4,-2) [draw,circle,minimum height=0.5cm] {};
  \end{tikzpicture}
  }
  %\subfloat[]{\includegraphics[width=38mm]{task_ims/clusters/basketball.png}}
  %\subfloat[]{\includegraphics[width=41mm]{ims/soccer_clustering.png}}
  \caption{\small{Example of an evolution forest associated to a single toy QD instance $Q_i$, initialized with the prior population $\{A, B, C\}$. Each edge indicates a parent-offspring relationship between two nodes. The $S_i$ are the solutions found by the QD algorithm. In this example, $f_0^i(A)=1, f_0^i(B)=2, f_0^i(C)=2$ and $f_1^i(A)=2, f_1^i(B)=3.5, f_1^i(C)=3$.}}
  %\textbf{(b)} 2d projection (using \cite{van2008visualizing}) of the weights of $3000$ successful policies on environments sampled from the \texttt{basket-v2} task distribution of the metaworld benchmark. The colors have been randomly associated to different clusters for better visualization. Those projections, which preserve neighbourhood information, seem to indicate that successful policies on this task do indeed form clusters. \textbf{(c)} The same as (b), but for the \texttt{soccer-v2} environment from that benchmark set.}}
  \label{fig_tree}
\end{figure}

\begin{algorithm}
  %\CommentSty{\color{blue}}
  \SetCommentSty{mycommfont}
  \begin{footnotesize}
    \KwIn{Train task distribution $\mathcal{T}_{train}$, sample size $M$ at each meta iteration, maximum number of meta iterations $G_{outer}$, prior population size $\mu$, number of offsprings $\lambda$, maximum number of optimization steps per problem $G_{QD}$, maximum number $s_{max}$ of successful policies per environment, constant factor $c_{\lambda}$ specifying population size in each QD optimization, mutation operator $\Delta$}
  \KwOut{Prior population $\mathcal{P}$}
  \ \\
  $\mathcal{P}=\texttt{InitializeRandomPopulation()}$ \\
  \SetAlgoLined
  \For{$g_o \in [0, G_{outer})$}
    {
      \texttt{tasks} $\leftarrow$ \texttt{SampleTasks}($\mathcal{T}_{train}$, $M$)\\
      \texttt{offsprings} $\leftarrow$ \texttt{$\Delta$}(\texttt{$\mathcal{P}$, $\lambda$}) \\
      \texttt{pop}=\texttt{offsprings} $\cup \mathcal{P}$ \\
      \For {$r \in $\texttt{pop}}
      {
        \tcp{counter of number of solutions that evolve from $r$}
        $r.\eta \leftarrow 0$ \\
        \tcp{list of depths of solutions in the evolution tree of $r$} 
        $r.\mathcal{D} \leftarrow \texttt{EmptyList()}$ \\
        \tcp{fitnesses to maximize (see equation \ref{eq_updates})}
        $r.\texttt{f}_0 \leftarrow 0 $ \\
        $r.\texttt{f}_1 \leftarrow -\infty$ \\ 
      }

      \For {$t_i \in$ \texttt{tasks}} 
      {
        \texttt{solutions}, \texttt{evolution\_paths} $\leftarrow$ \texttt{QualityDiversityAlgorithm}($t_i$, $\psi_{i}$, \texttt{pop}, $c_{\lambda}$, $s_{max}$)\\
        \For {$s \in$ \texttt{solutions}}
        {
          \tcp{get root of corresponding evolution tree}
          $r \leftarrow$ \texttt{GetRoot($s$, \texttt{evolution\_paths})} \\
          $r.\eta \leftarrow r.\eta + 1$\\
          $r.\mathcal{D} \leftarrow \texttt{concatenate(}r.\mathcal{D},\{s.$\texttt{depth}$\}$\texttt{)} \\
        }
      }
      \For {$r \in $\texttt{pop}}
      {
        \If {$r.\eta \neq 0$}
        {
          $r.\texttt{f}_0 \leftarrow r.\eta$ \\
          $r.\texttt{f}_1 \leftarrow -1 \times$ \texttt{mean}($r.\mathcal{D}$) \\
        }
      }

      \tcp{\textit{e.g.} with NSGA2}
      $\mathcal{P} \leftarrow $ \texttt{ParetoBasedSelection}(\texttt{pop}, $\texttt{f}_0$, $\texttt{f}_1$)
    }
  \caption{\small{The FAERY algorithm}}
  \label{algo_pseudo}
  \end{footnotesize}
\end{algorithm}

\section{Experiments}
\label{sec_experim}

We first provide formal definitions and outline the objectives of the experiments.

\textbf{Single-task vs cross-task generalization.} In this section, we will consider that each task $t_i$ has a set of goal areas $\mathcal{H}(t_i)=\{h_1^i, ..., h_s^i\}$ in some Euclidean \textit{goal space}. The objects of interest in the environment (\textit{e.g.} objects to assemble in a robot manipulation task, or walls in a maze) will be noted $\Upsilon(t_i)=\{o_1^i, ..., o_{\upsilon}^i\}$.

Let $\mathcal{T}$ denote a task distribution. We will consider that $\mathcal{T}$ is a \textit{single-task distribution} iif the goals and initial object poses of any two environments $t_i, t_j \sim \mathcal{T}$ (with $i\neq j$) are related by Euclidean transformations: $\forall h^i \in \mathcal{H}(t_i), \forall o^i \in \Upsilon(t_i)$, there exists a unique $h^j \in \mathcal{H}(t_j)$ and a unique $o^j \in \Upsilon(t_j)$ such that $h^j=R_{h}^{ij}(h^i)$ and $o^j=R_o^{ij}(o^i)$, where $R_{h}^{ij}, R_{o}^{ij}$ are some Euclidean transforms. An example of such task distributions is a robot manipulation task where the aim is to grasp a ball and throw it in a basket, and where the poses of the ball and the basket are randomly initialized. We will say about experiments evaluating the generalization capacities of an algorithm with such a $\mathcal{T}$ that they assess the \textit{single-task} or \textit{intra-task} generalization capabilities of that algorithm. In such cases, it seems reasonable to expect some degree of generalization, \textit{i.e.} that knowledge acquired in some $t_i$ is likely to help in solving some $t_j$ from that distribution in the future.

Now, consider a task distribution $\mathcal{T}_{manip}$ which is used to randomly assign dramatically different tasks to a robotic arm. For example, $t_i, t_j \sim \mathcal{T}_{manip}$ could respectively correspond to assembling lego pieces and hitting a ball so that it goes through a hole. In that case, it is not immediately obvious whether knowledge gained in one environment will facilitate learning in the other. The corresponding POMDPs could greatly differ. We will refer to generalization in such a setting as \textit{cross-task} generalization. In this work, we restrict ourselves to cross-task transfer between POMDP distributions which share the same action set, and whose state-spaces have the same dimensionality (without necessarily overlapping). 

Note that in contrast with many works on transfer learning, we are not interested in transfer between two different fixed POMDPs, but between two POMDP distributions.

\textbf{Experiments outline.} While FAERY is formulated with both intra-task and cross-task generalization in mind, we will primarily focus on evaluating its use in few-shot adaptation for single-task generalization (\S\ref{experiments_single}). Regarding the cross-task setting, we only provide cross-task knowledge transfer results based on priors learned on a source distribution (\S\ref{ctg_sec}). The reason for this is that large-scale experiments in the cross-task setting are costly, and so we leave exhaustive investigations into that matter for future works. 

In all experiments, we used NSGA2 \cite{deb2002fast} with Novelty \footnote{For the reader's convenience, the definition of Novely is stated in this footnote. Consider a policy $\theta$ and its behavior descriptor $b_{\theta}$. Let $R_{ref}\triangleq\{\theta_j\}_j$ be a reference set (usually composed of the current population and a subset of previously visited individuals). Then the Novelty score of $\theta$ is computed as $\frac{1}{K}\sum_{i=0}^K d(b_{\theta},b_{\theta_i})$ where the $\{\theta_i\}_{i=0}^K$ are such that the corresponding $b_{\theta_i}$ are the $K$ nearest neighbors of $b_{\theta}$ in $R_{ref}$.} \cite{lehman2011abandoning} and fitness objectives as the underlying QD method, in place of the \texttt{QualityDiversityAlgorithm} placeholder from algorithm \ref{algo_pseudo}. As the data-efficiency issues that are addressed by FAERY stem from evolutionary mechanisms that are common to all QD families, it is reasonable to expect that results obtained with NSGA2 will also be a strong indicator of what can be expected of the application of FAERY to other QD algorithms.

\begin{figure}[ht!]
  \centering
  \vspace*{0.1cm}
  \hspace*{0.6cm}
  \includegraphics[width=24mm]{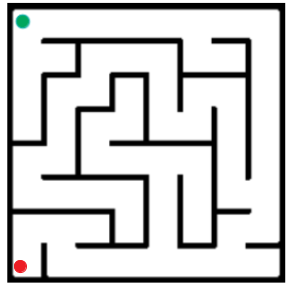}
  \hspace*{0.58cm}
  \includegraphics[width=24.4mm]{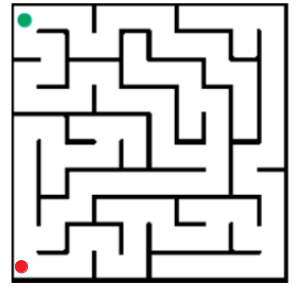}\\
  \includegraphics[width=32mm,trim={0 0 0 1.2cm},clip]{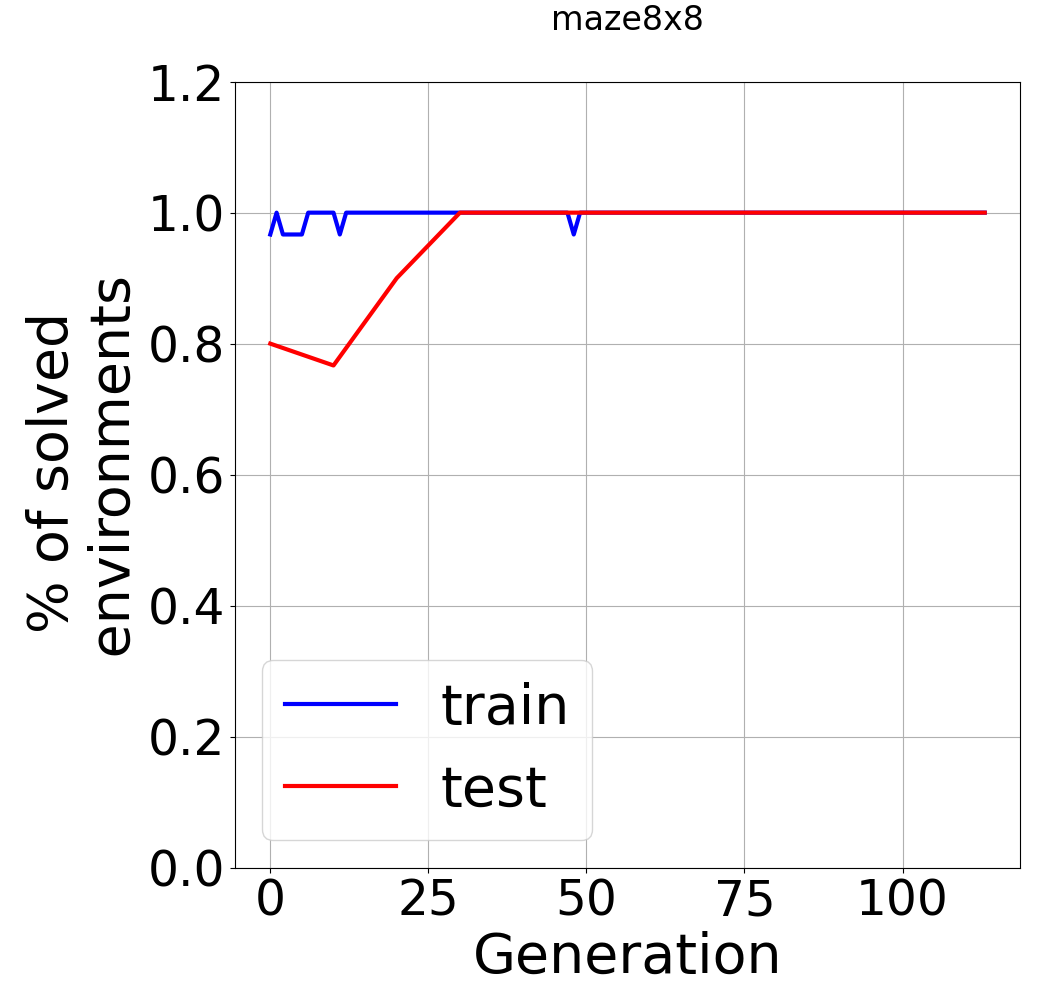}
  \includegraphics[width=32mm,trim={0 0 0 1.2cm},clip]{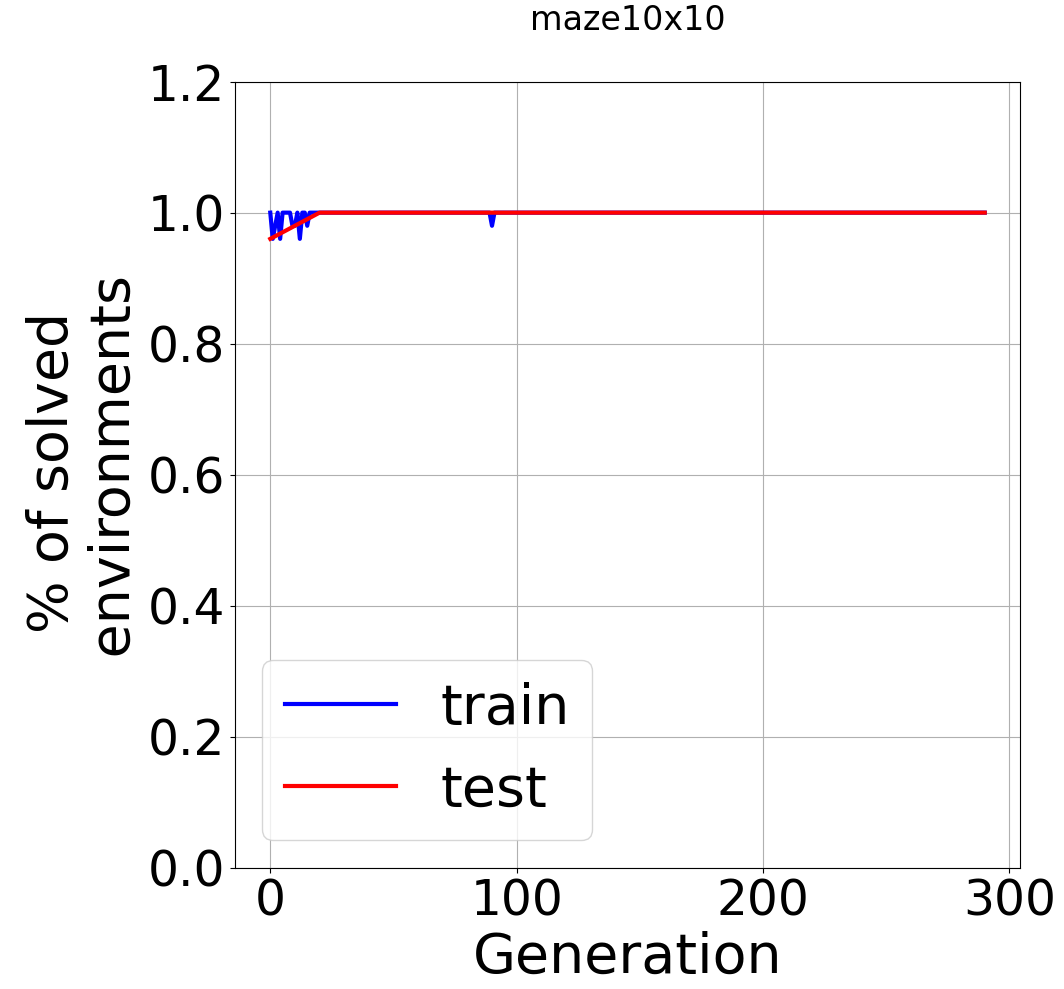}\\
  \includegraphics[width=32mm,trim={0 0 0 1.2cm},clip]{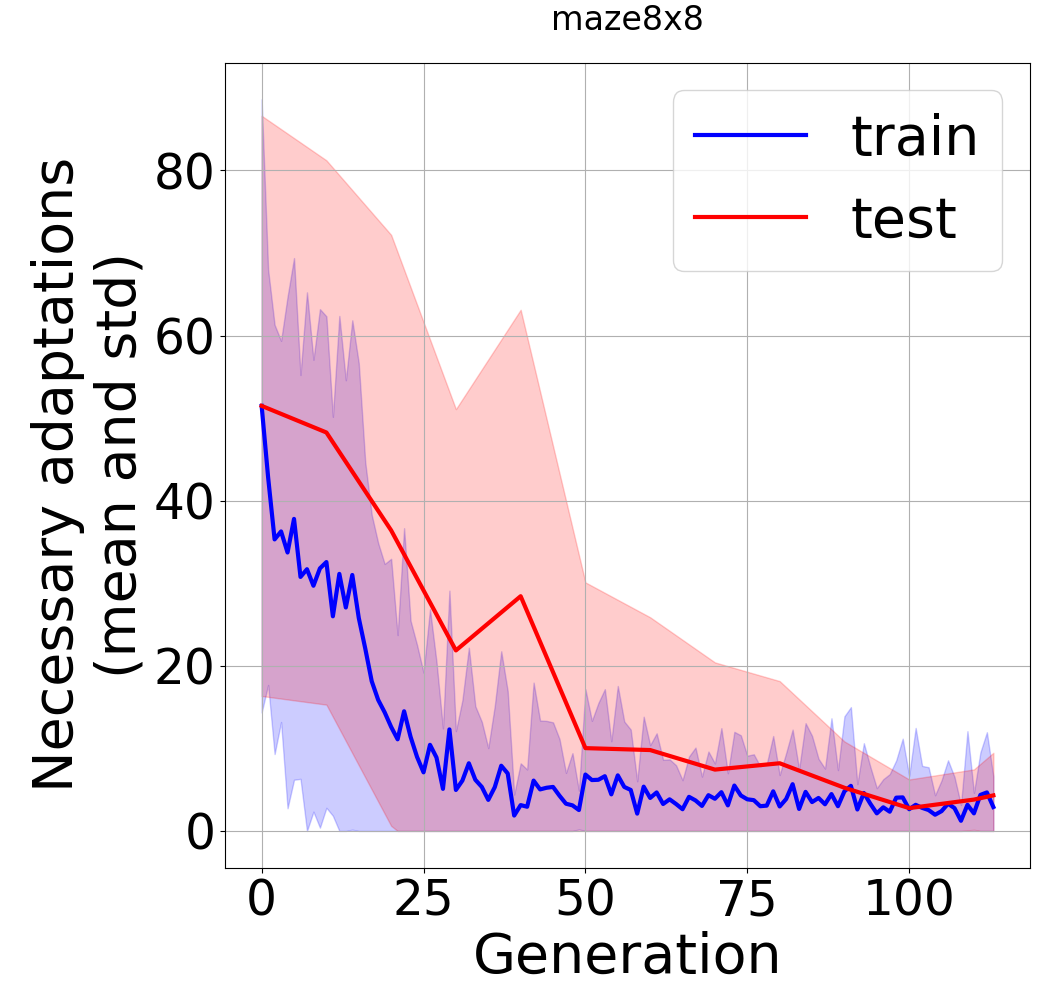}
  \includegraphics[width=32mm,trim={0 0 0 1.2cm},clip]{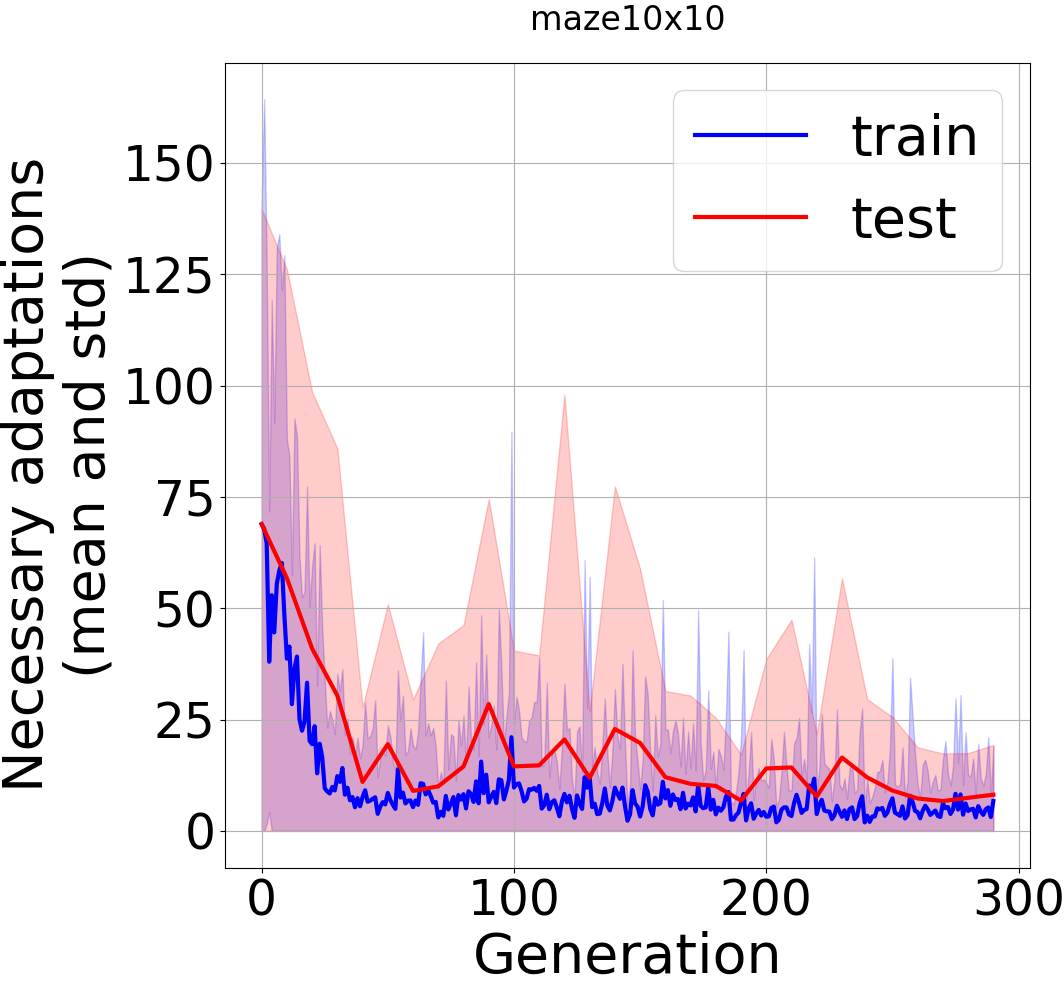}
  \caption{\small{Examples of random mazes sampled from the maze distributions, and the results of the proposed method (second and third rows). The first and second columns respectively correspond to $8\times 8$ and $10\times 10$ mazes. The horizontal axis of the plots in the second and last rows indicates the number of generations for which the prior population has been optimized by FAERY (\textit{i.e.} the number of meta-updates), which should not be confused with the number of generations needed for a single QD algorithm to converge. Notice that on this axis, generation $0$ corresponds to a QD optimization that is performed from scratch as no priors have been learned up to this point. Note that all train/test environments that are sampled at generation $i$ on this axis will have their corresponding QD instances initialized using the priors that are obtained from generation $i-1$. \textbf{(Top row)} Example mazes that are sampled from the $8 \times 8$ and $10 \times 10$ maze distributions. \textbf{(Middle row)} The ratio of environments among the M sampled ones that the QD instances are able to solve. \textbf{(Bottom row)} the average number of generations necessary to solve the tasks (that are solvable at a given generation).}}
    \label{mazes_fig}
\end{figure}

The policies are implemented as fully connected feed-forward neural networks with $tanh$ activations and the bounded polynomial operator \cite{deb2002fast} is used for mutations. As recommended in \cite{gomes2015devising}, the Novelty objective is computed based on the $k=15$ nearest neighbors, using a maximum archive size of $5000$ individuals. We will note $M_{train}, M_{test}$ the number of training and testing environments. Note that train/test datasets are disjoint in all the experiments, \textit{i.e.} an environment sampled from the test distribution $\mathcal{T}_{test}$ has zero probability of being sampled from $\mathcal{T}_{train}$ and vice versa.

\begin{figure}[ht!]
    \centering
    \vspace*{0.1cm}
    \hspace*{0.4cm}
    \includegraphics[width=25mm]{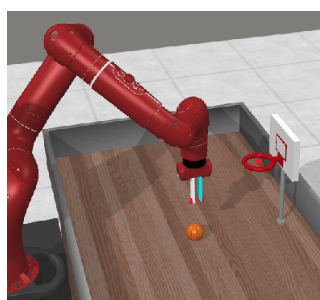}
    \hspace*{0.7cm}
    \includegraphics[width=25mm]{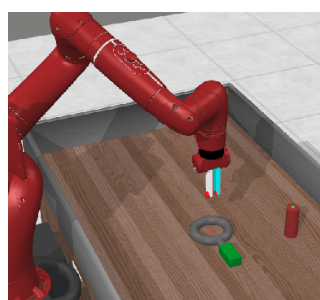}\\
    \includegraphics[width=32mm,trim={0 0 0 1.2cm},clip]{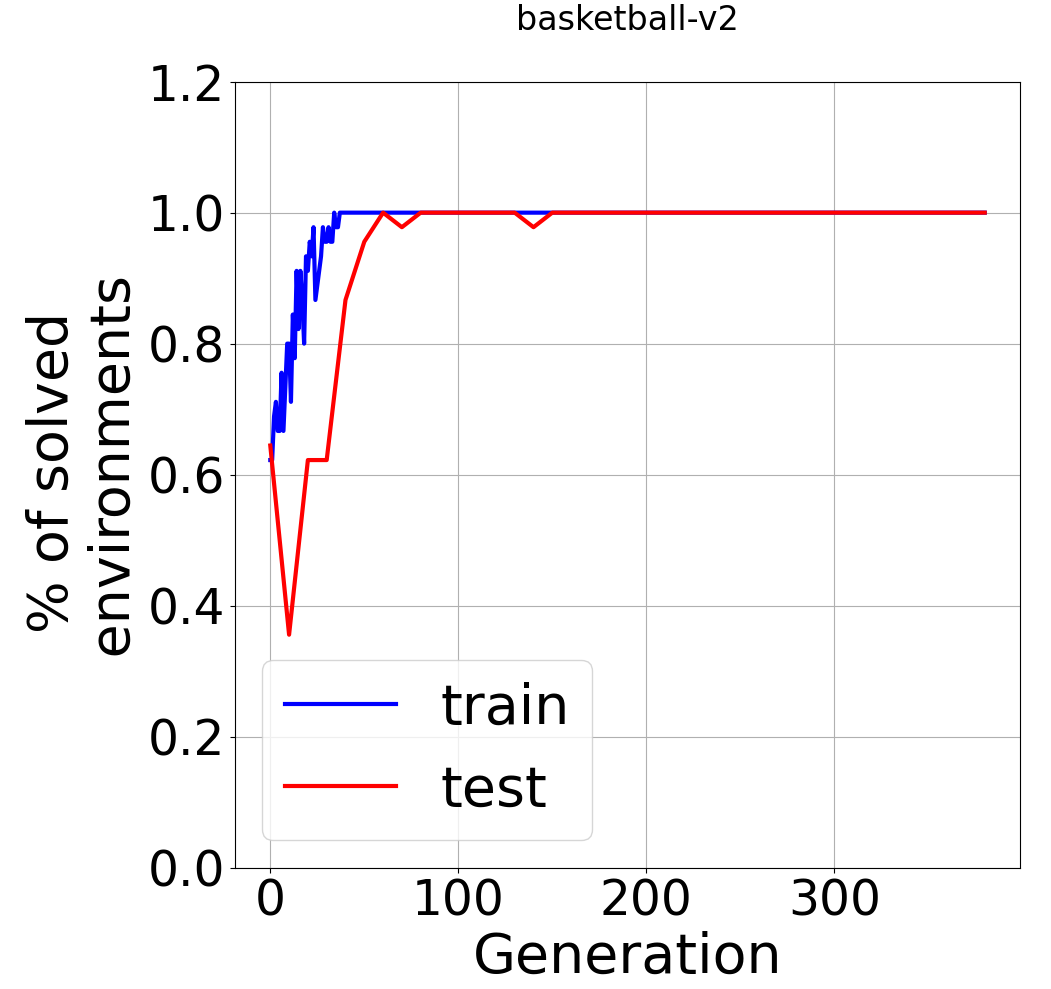}
    \includegraphics[width=32mm,trim={0 0 0 1.2cm},clip]{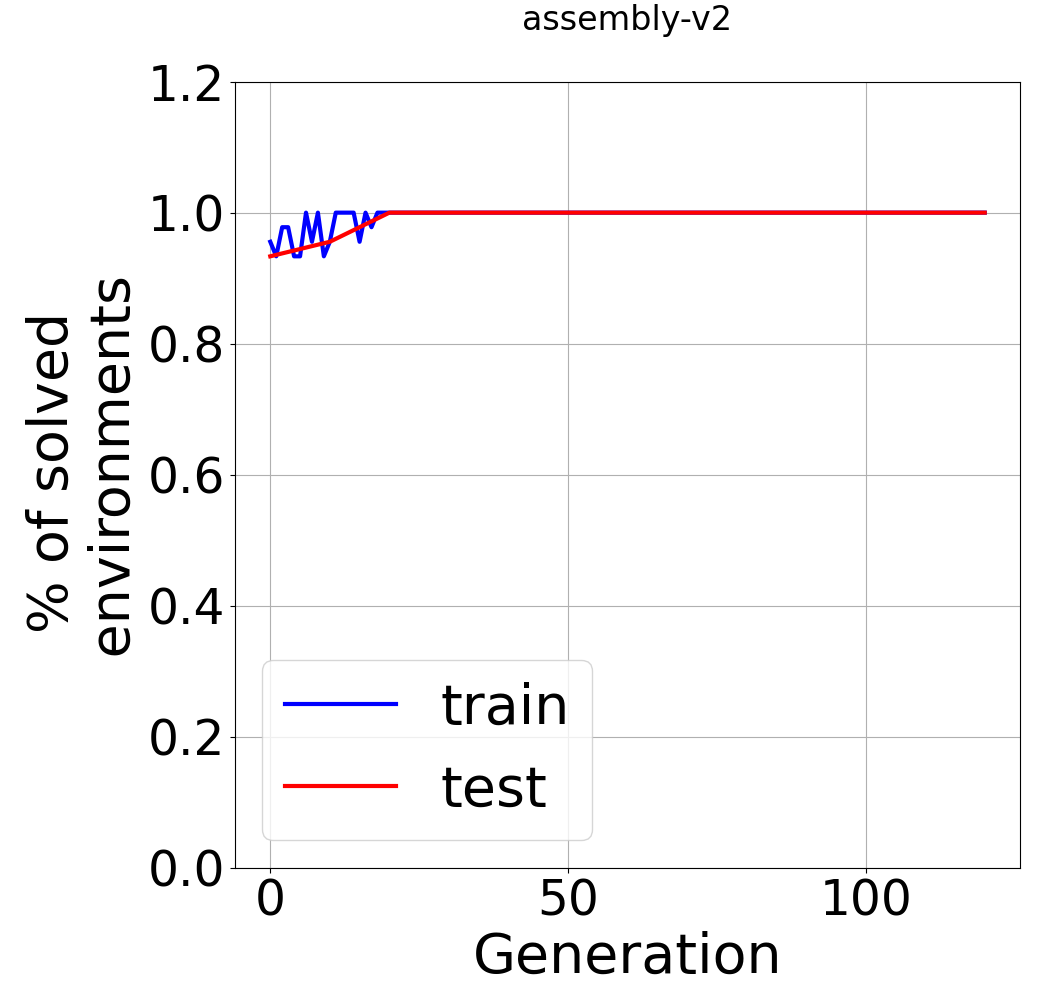}\\
    \includegraphics[width=32mm,trim={0 0 0 1.2cm},clip]{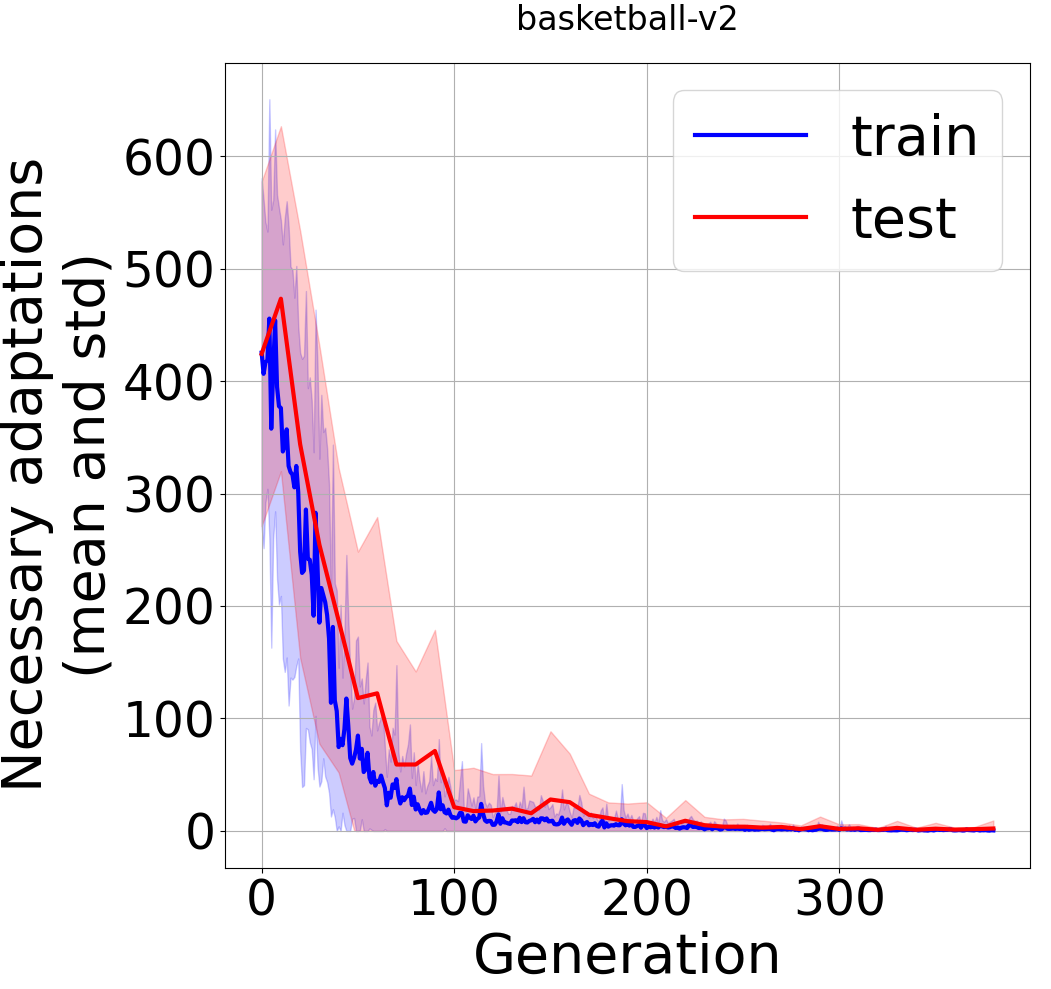}
    \includegraphics[width=32mm,trim={0 0 0 1.2cm},clip]{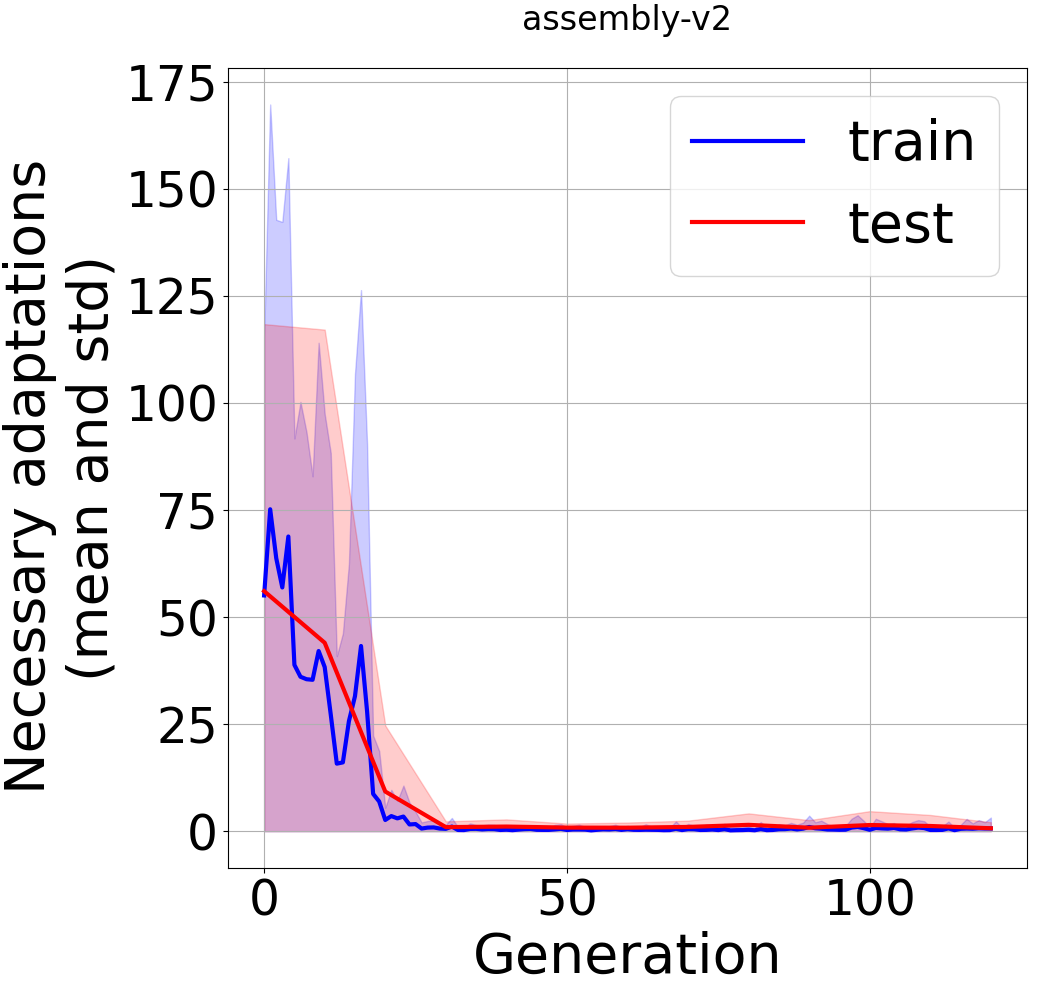}\\
    %\subfloat[]{\includegraphics[width=120mm]{task_ims/complete_figs_2/disassemble.png}}\\
    \caption{\small{Each of the two columns corresponds to an example task from metaworld. \textbf{(Top row)} Example frames from the \texttt{basketball-v2} and \texttt{assembly-v2} tasks from the benchmark. \textbf{(Middle row)} The ratio of environments among the $M$ sampled ones that the QD instances are able to solve at each generation of FAERY. For example, on the \texttt{basketball-v2} task, we see that without optimized priors (generation 0 on the horizontal axis), only $60\%$ of the $M$ QD methods are successful. After about $100$ meta-updates to the prior population however, we see that that all QD methods successfully solve their environments. \textbf{(Bottom row)} The average number of generations necessary to solve tasks (that are solvable at a given generation). As in the maze experiments, the horizontal axis of the plots is the number of generations for which the prior population is optimized by FAERY (\textit{i.e.} the number of meta-updates), which should not be confused with the number of generations needed for a single QD algorithm to converge. Note that all train/test environments that are sampled at generation $i$ on this axis will have their corresponding QD instances initialized using the priors that are obtained from generation $i-1$.}}
    \label{intratask_examples}
\end{figure}

\subsection{Few-shot Learning For Single-Task Generalization}
\label{experiments_single}

Our aim in this section is to demonstrate for the single-task setting, in both sparse and dense reward scenarios, that the proposed algorithm reduces the number of generations necessary for QD algorithms to obtain a set of diverse but well-performing agents. We performed experiments in those two environments:

\begin{itemize}
  \item Navigation tasks in randomly generated mazes composed of $8\times8$ and $10\times10$ cells. 
  \item Robotic manipulation environments from the meta-world \cite{yu2020meta} benchmark, which is based on the mujoco physics engine\footnote{\url{http://www.mujoco.org/}}.
\end{itemize}

The mazes differ from each other in the placement of walls and thus in layout, and the manipulation tasks differ in the poses of the goals and initial placements of objects of interest.

\subsubsection{Few-Shot Maze Navigation}

For this experiment, two datasets of randomly generated mazes\footnote{Maze generation is a classical topic in procedural content generation \cite{de2019procedural}. We used a simple depth-first search approach with back-tracking.} were used. The first one, that we will designate by \texttt{mazes8x8} is comprised of $1200$ train and $200$ test mazes, where the wall positions have been sampled from an $8\times8$ grid. The second dataset, \texttt{mazes10x10} is comprised of $2400$ train and $400$ test mazes, and was similarly sampled based on a $10\times10$ grid. Examples from those environments are given in the top row of figure \ref{mazes_fig}. In all mazes, the agent starts in the area in the bottom left marked with a red circle, and the goal of the environment is to reach the top left area marked with a green circle. What changes from an environment to another is the layout of the maze. In other terms, the set $\Upsilon$ is composed of cell borders, which constitute building blocks for walls.

%\begin{figure}[H]
%    \centering
%    \hspace*{-1.0cm}\subfloat[]{\includegraphics[width=80mm]{task_ims/complete_figs/basket.png}}
%    \subfloat[]{\includegraphics[width=80mm]{task_ims/complete_figs/assembly.png}}\\
%    \hspace*{-1.0cm}\subfloat[]{\includegraphics[width=80mm]{task_ims/complete_figs/hammer.png}}
%    \subfloat[]{\includegraphics[width=80mm]{task_ims/complete_figs/buttonpress_topdown.png}}\\
%    \hspace*{-1.0cm}\subfloat[]{\includegraphics[width=82mm]{task_ims/complete_figs/plateslideback.png}}
%    \subfloat[]{\includegraphics[width=80mm]{task_ims/complete_figs/soccer.png}}\\
%    \hspace*{-1.0cm}\subfloat[]{\includegraphics[width=82mm]{task_ims/complete_figs/handinsert.png}}
%    \subfloat[]{\includegraphics[width=80mm]{task_ims/complete_figs/disassemble.png}}\\
%    \caption{}
%    \label{intratask_examples}
%\end{figure}
%
 The perception and actions of the agent are similar to those that were introduced in the deceptive maze example from \cite{lehman2011abandoning}. The agent is equipped with proximity sensors which allow it to sense the presence of walls\footnote{More precisely, two Boolean "bumpers" that detect collisions with walls and three rangefinders which are each associated to a single direction. Taking as reference the axis pointing in the forward direction of the robot, the latter are positioned at angles $-45$\textdegree, $0$\textdegree , $45$\textdegree\ and return the distance to the closest intersection point along those rays. Their range is limited to about one tenth of the environment.}. The observation space is $5$-dimensional, and the action space is the $2-d$ force vector. The rewards in this environment are sparse: the agent only receives a positive reward when the goal area is reached. As a result, QD optimization in such an environment often behaves in a manner that is equivalent to pure Novelty search.

The shallow networks that we used to encode the policies had $3$ hidden layers, each of dimension $10$. For \texttt{mazes8x8}, we set $\lambda=\mu=24$, and the number of sampled train mazes in each iteration of FAERY was set to $M_{train}=M_{test}=30$. In the \texttt{mazes10x10} experiments, both population and offspring sizes where set to $40$, and we had $M_{train}=M_{test}=50$. As in the original paper that introduced the Novely metric \cite{lehman2011abandoning}, we used the final $2d$ position of the robot in a trajectory as its behavior descriptor for computing Novelty. 

\begin{table*}[ht]
\begin{notsotiny}
  \vspace*{0.18cm}
  \begin{threeparttable}
  \begin{tabularx}{0.98\textwidth} {bssmmmsttt}
 \hline
    task name & QD $\mathcal{R}^{\text[1]}$ & QD $\mathcal{G}^{\text[2]}$ & FAERY $\mathcal{R}^{\text[3]}$ & FAERY $\mathcal{G}^{\text[4]}$ & FAERY $it_c^{\text[5]}$ & $\mu^{\text[6]}$ & $\lambda^{\text[7]}$ & $M_{train}^{\text[8]}$ & $M_{test}^{\text[9]}$ \\
 \hline
    assembly-v2 & \textcolor{red}{0.88}  & \textcolor{red}{72.6}  & \textbf{1.0}  & \textbf{0.68} & 35 & 35 & 35 & 45 & 45 \\
 \hline
    door-close-v2 & 1.0 & 1.3 & 1.0 & \textbf{0.1} & 15 & 35 & 35 & 45 & 45 \\
 \hline
    window-open-v2 & 1.0 & 0.26 & 1.0 & \textbf{0.0} & 15 & 130 & 130 & 50 & 40 \\
 \hline
    drawer-open-v2 & 1.0 & \textcolor{red}{9.75}  & 1.0 & \textbf{1.93} & 250 & 35 & 35 & 45 & 45 \\
 \hline
    sweep-into-v2 & 1.0 & \textcolor{red}{15.28} & 1.0  & \textbf{3.6} & 100 &  80 & 80 & 25 & 25 \\
 \hline
    basketball-v2 & \textcolor{red}{0.63} & \textcolor{red}{449.0} & \textbf{1.0} & \textbf{1.22} & 275 & 40 & 40 & 45 & 45 \\
 \hline
    button-press-wall-v2 & 1.0 & 3.2 & 1.0 & \textbf{0.1} & 15 & 35 & 35 & 45 & 45 \\
 \hline
    door-lock-v2 & 1.0 & \textcolor{red}{30.63} & 1.0 & \textbf{0.4} & 290 & 40 & 40 & 45 & 45 \\
 \hline
    handle-pull-v2 & 1.0 & \textcolor{red}{6.6} & 1.0 & \textbf{0.62} & 100 &  35 & 35 & 25 & 25 \\
 \hline
    disassemble-v2 & 1.0 & \textcolor{red}{34.13} & 1.0 & \textbf{0.35} & 150 & 35 & 35 & 45 & 45 \\
 \hline
    sweep-v2 & 1.0 & \textcolor{red}{32.18} & 1.0 & \textbf{1.38} & 35 & 130 & 130 & 50 & 40 \\
 \hline
    box-close-v2 & 1.0 & 3.5 & 1.0 & \textbf{0.44} & 100 &  130 & 130 & 30 & 25 \\
 \hline
    dial-turn-v2 & 1.0 & 1.82 & 1.0 & \textbf{0.04} & 10 & 35 & 35 & 45 & 45 \\
 \hline
    hand-insert-v2 & \textcolor{red}{0.91} & \textcolor{red}{29.87} & \textbf{1.0} & \textbf{2.0} & 150 & 35 & 35 & 45 & 45 \\
 \hline
    soccer-v2 & 1.0 & 1.2 & 1.0 & \textbf{0.0} & 10 & 40 & 40 & 20 & 20 \\
 \hline
    button-press-topdown-v2 & \textcolor{red}{0.62} & \textcolor{red}{203.17} & \textbf{1.0} & \textbf{0.02} & 210 & 35 & 35 & 45 & 45 \\
 \hline
    push-v2 & 1.0 & \textcolor{red}{9.7} & 1.0 & \textbf{0.4} & 100 &  100 & 100 & 10 & 10 \\
 \hline
    stick-push-v2 & 1.0 & \textcolor{red}{24.42} & 1.0 & \textbf{0.02} & 80 & 40 & 40 & 45 & 45 \\
 \hline
    window-close-v2 & 1.0 & 2.6 & 1.0 & \textbf{0.0} & 10 & 40 & 40 & 45 & 45 \\
 \hline
    reach-wall-v2 & 1.0 & 1.75 & 1.0 & \textbf{0.37} & 15  & 40 & 40 & 45 & 45 \\
 \hline
    plate-slide-v2 & 1.0 & 2.7 & 1.0 & \textbf{0.0} & 10 & 100 & 100 & 10 & 10\\
 \hline
    plate-slide-back-v2 & 1.0 & \textcolor{red}{123.73} & 1.0 & \textbf{0.4} & 90 & 40 & 40 & 45 & 45 \\
 \hline
    plate-slide-back-side-v2 & 1.0  & \textcolor{red}{132.4} & 1.0 & \textbf{0.02} & 90 & 40 & 40 & 45 & 45 \\
 \hline
    hammer-v2 & \textcolor{red}{0.94} & \textcolor{red}{81.58} & \textbf{1.0} & \textbf{5.2} & 120 & 40 & 40 & 45 & 45 \\
 \hline
    lever-pull-v2 & 1.0 & 3.9 & 1.0 & \textbf{0.0} & 80 & 100 & 100 & 10 & 10\\
 \hline
  \end{tabularx}
    \begin{tablenotes}
    \item[1] \textbf{QD $\mathcal{R}$}: percentage of solved environments (normalized in $[0,1]$) when using QD optimization from scratch.
    \item[2] \textbf{QD $\mathcal{G}$}: Average number of necessary generations required to solve an environment, when using QD optimization from scratch.
    \item[3] \textbf{FAERY $\mathcal{R}$}: percentage of solved environments (normalized in $[0,1]$) when using QD optimization with priors learned by FAERY.
    \item[4] \textbf{FAERY $\mathcal{G}$}: Average number of necessary generations required to solve an environment, when using QD with priors learned by FAERY.
    \item[5] \textbf{FAERY $it_c$}: Number of generation at which the prior population has converged in terms of FAERY $\mathcal{R}$ and FAERY $\mathcal{G}$.
    \item[6,7] \textbf{$\mu, \lambda$}: respectively population size and number of offsprings.
    \item[8,9] \textbf{$M_{train}, M_{test}$}: number of environments used for training and testing.
    \end{tablenotes}
  \end{threeparttable}
  \caption{\small{Comparison of ratio of solved environments and adaptation speed between QD optimization from scratch and QD optimization based on FAERY priors. For each task, the results are averaged on 3 different realizations of samples of size $M_{test}$. The entries that are highlighted using the color red are those where QD optimization from scratch either fails to solve all environments or takes $>20$ generations on average to solve an environment. Values corresponding to tasks where the proposed method results in significant improvements are highlighted with bold fonts. Note that the values of $\mu, \lambda$ and $M_{test}, M_{train}$ were mainly dictated by our available computational resources at the time of the experiments, and are only reported for reproducibility.}}
\label{table_singetask}
\end{notsotiny}
\end{table*}
 
 The results for both cases are reported in figure \ref{mazes_fig}. In both experiments, we can see that initially, the $M$ instance of QD optimization that are initialized with a random population i) Are not able to solve all randomly sampled tasks (middle row of figure \ref{mazes_fig}) and ii) take many generations (in average, $\sim 60$ and $\sim 75$ for \texttt{mazes8x8} and \texttt{mazes10x10} respectively) to solve the sampled environments that they are able to solve. However, once FAERY starts learning the prior population $\mathcal{P}$, we see that the ratio of solved environments at each generation increases to $100\%$, and the number of required generations dramatically decreases during both training and testing, until it stabilizes at a much smaller average ($<5$ for the \texttt{mazes8x8} and $\sim 8$ for \texttt{mazes10x10}). Note that the bottom row of figure \ref{mazes_fig} excludes the environments that have not been solved, which means that it understates the number of optimization steps that are wasted by QD optimizations that do not use the priors learned by FAERY. For example, in the very first iteration of the \texttt{mazes8x8} experiments, only about $80\%$ of the environments are solved, which means that the remaining $20\%$ of the QD instances run for the maximum allowed number of generations $G_{QD}$, which was set to $200$ in this experiment.

\subsubsection{Few-shot Single-Task Robotic Manipulation}

The networks that we used to encode the policies for these experiments have a single hidden layer of dimension $50$. The values of $\mu, \lambda$ and $M_{test}, M_{train}$ were set on a per-task distribution basis, and were mainly dictated by our available computational resources at the time of the experiments. To accelerate overall QD optimization, we used the dense rewards that have been made available in recent releases of the benchmark (v2). The behavior descriptors that we used to compute the novelty objective were the final $3d$ positions of the objects of interest being manipulated: for example, in the \texttt{basketball-v2} task, this was the final position of the ball, and in the \texttt{assembly-v2} task, this was the $6d$ concatenation of the final positions of the two objects being assembled. Note that each task name such as for example \texttt{hammer-v2} designates a distribution of tasks, from which $M$ environments can be sampled, that will differ in terms of initial conditions of objects and/or goals. For all experiments in this section, $G_{QD}$ was set to $800$.

We performed experiments on $25$ task distributions chosen from the ones available in the subset of metaworld \cite{yu2020meta} dedicated to single-task generalization (the ML1 benchmark). The $25$ tasks were chosen according to the following criteria and constraints: first, meta-learning experiments in general become very demanding in terms of computational resources when the problems being solved are based on physical simulations. It was thus impractical for us to consider all environments. Second, many environments in the metaworld benchmark, especially those where the task is correlated with simple (\textit{e.g.} $1d$) behavior descriptors can readily be solved by QD methods in very few generations. While we did include some of these tasks in our experiments, we tried to focus on environments that were more challenging for QD methods. Finally, we observed that the solutions found by QD optimization in some environments such as \texttt{bin-picking-v2} and \texttt{shelf-place-v2} are very often brittle as they exploit particularities/bugs in the physics engine to achieve the goal. While reasonable, generalizable behaviors eventually emerge, they are rare enough that they make the number of generations required for QD optimization from scratch prohibitive. Therefore, we do not include them in this subsection, but will use them in the next one \ref{ctg_sec} to show the potential of FAERY in providing "stepping stones" \cite{wang2019paired} for generalization across tasks. 

Figure \ref{intratask_examples} illustrates the performance of our method on two of the selected environments. As in the navigation experiments, the middle rows of this figure indicate that QD methods for which no priors are available (generation 0) are not able to solve all of the sampled environments. This effect seems to be much more pronounced for \texttt{basketball-v2}, where a QD method initialized from scratch seems incapable of solving more than $60\%$ of the sampled environments. However, as FAERY learns a suitable prior $\mathcal{P}$, the number of solved environments sampled from those tasks rises to a $100\%$. Similarly, the number of generations required for QD optimizations significantly decreases: for both of those tasks, it falls from hundreds of generations to $<3$ on average.

More detailed results on all $25$ environments are presented in table \ref{table_singetask}. For simplicity in the tabulation of the results, we define the following symbols: $\mathcal{R}$ will denote the ratio of solved environments among the $M$ environments that are considered at anyone time. The average number of necessary generations is noted $\mathcal{G}$. Note that in that table QD $\mathcal{R}$, QD $\mathcal{G}$ correspond to QD optimization on $M$ test problems from scratch, and FAERY $\mathcal{R}$, FAERY $\mathcal{G}$ correspond to QD optimization using the learned priors after convergence. In the table, the generation number at which the priors have converged is given in the column titled FAERY $it_c$. For reproducibility, parameters of the experiments ($\lambda, \mu, M_{train}, M_{test}$) are also reported. In the table, the QD $\mathcal{R}$ and QD $\mathcal{G}$ entries of the tasks where QD optimization from scratch fails to solve all environments or takes $>20$ generations on average are highlighted with the color red. The FAERY $\mathcal{R}$, FAERY $\mathcal{G}$ values of Tasks where the priors learned by FAERY result in significant improvements are written using bold fonts. As in the previously discussed examples, it can be seen that QD optimization from scratch is often unable to solve all sampled environments and requires many generations before finding a solution. Both of these aspects significantly improve once our method starts learning the prior population: at convergence, all newly sampled environments are solved in few (generally $<3$) generations.\\
\textbf{Time savings.} The computational speedup when using the learned priors on a new environment results from the reduction in the number of necessary adaptations. For a maximum computational budget of $G_{QD}$ generations and noting $time(\Delta)$ the time taken by each QD population update, an approximate lower bound to the time savings is given by $({\text{QD}\mathcal{R}\times \text{QD}\mathcal{G} + (1-\text{QD}\mathcal{R})\times\text{G}_\mathcal{QD}}-{\text{FAERY}_{G}})\times time(\Delta)$, where the notations are those that were defined in table \ref{table_singetask}. In our experiments on metaworld, a single update step of the population takes on average $\sim 1.8$ seconds for population sizes of $35$ and $40$, and takes $\sim 2$ seconds for population sizes of $100$ and above when running on $48$ Intel Xeon Silver 4214 CPU cores. As a result, in tasks such as \texttt{basketball-v2}, using the priors enables adaptation in a matter of seconds where learning from scratch would take $\sim 15$ to $20$ minutes while leaving some environments unsolved. Note however that in a few tasks such as \texttt{box-close-v2}, only a few seconds will be saved by using the priors in a new environment as $\text{QD}\mathcal{G}$ is already low. Nonetheless, even in such cases, learning and using the priors can be advantageous due to the cumulated time that will be saved on (potentially numerous) future problems, especially in applications such as continual learning or those that require online/near real-time performance on new problems.
%\st{In environments where the advantage of using the priors is less pronounced, the cost of learning the priors themselves is also low or negligible, as indicated by the FAERY $it_{c}$ column in table I.}%\ref{table_singetask}.

\begin{table}[]
\begin{notsotiny}
  \begin{tabularx}{\columnwidth} {vtvv}
 \hline
    task name & prior type & \% envs solved & Mean required QD generations \\
 \Xhline{3\arrayrulewidth}
    shelf-place-v2 & None & 0.0 & N.A \\
 \hline
    shelf-place-v2 & basketball-v2 solvers & 20\% & 582 \\
 \hline
    shelf-place-v2 & basketball-v2 FAERY & \textbf{24\%} & 650.25  \\
 \Xhline{3\arrayrulewidth}
    bin-picking-v2 & None & 95\% & 130.68  \\
 \hline
    bin-picking-v2 & basketball-v2 solvers & 93\% & 183  \\
 \hline
    bin-picking-v2 & basketball-v2 FAERY & \textbf{100\%} & 151  \\
 \Xhline{3\arrayrulewidth}
    pick-place-v2 & None & 51\% & 117 \\
 \hline
    pick-place-v2 & basketball-v2 solvers & 55\% & 90  \\
 \hline
    pick-place-v2 & basketball-v2 FAERY & \textbf{77\%} & 119  \\
 \Xhline{3\arrayrulewidth}
\end{tabularx}
  \caption{\small{\textbf{Cross-task generalization.} Results of QD optimization (NSGA2 with Novelty and fitness) averaged over $135$ environments on three different tasks with different prior types. This results seem to hint that using the priors learned by the proposed method can result in significant boosts in terms of percentage of solved environments. Note that the last column indicates the average number of generations per \textit{solved} environment.}, which understates the gains in terms of time savings as QD instances that are unable to solve an environment will run for $G_{QD}$ generations.}
  \label{table_ctg}
\end{notsotiny}
\end{table}
 
\subsection{Cross-Task Generalization}
\label{ctg_sec}

The objective of this experiment is to evaluate the potential use of the priors learned by FAERY in knowledge transfer from a source task $\mathcal{T}_{source}$ to target distributions $\{\mathcal{T}_{target}^i\}_i$ in the cross-task settings discussed in the beginning of section \ref{sec_experim}.

Assuming that we are given a target distribution $\mathcal{T}_{target}^i$ and an appropriate source task $\mathcal{T}_{source}$, the question of how to transfer knowledge from the latter to an agent learning on the former remains. Many approaches, such as MDP homomorphism based ones \cite{castro2010using, sorg2009transfer} rely on learning explicit mappings between fixed MDPs. As our aim is transfer learning from a distribution $\mathcal{T}_{source}$ rather than from a particular (PO)MDP, learning such a mapping is impractical. Furthermore, as previously stated in section \ref{sec_experim}, we are interested in cross-task transfer learning between POMDP distributions that share their action spaces, and whose state spaces have the same dimensionality. Therefore, a suitable transfer mechanism in this setting is to simply fine-tune a population of policies learned on environments from $\mathcal{T}_{source}$ on environments from $\mathcal{T}_{target}$. A way to do this would be to take a set of policies $Z=\{\theta_1, ..., \theta_{\mu}\}$ that solve environments $\{t_1, ..., t_{\mu}\}$ randomly sampled from $\mathcal{T}_{source}$. In this section, we show that using the priors $\mathcal{P}$ learned by FAERY is more advantageous than using $Z$.

\begin{figure}[h]
  \centering
    \subfloat[]
    {
      \begin{tikzpicture}
        \node[anchor=south west,inner sep=0] (image) at (0,0)
        {\includegraphics[width=25mm]{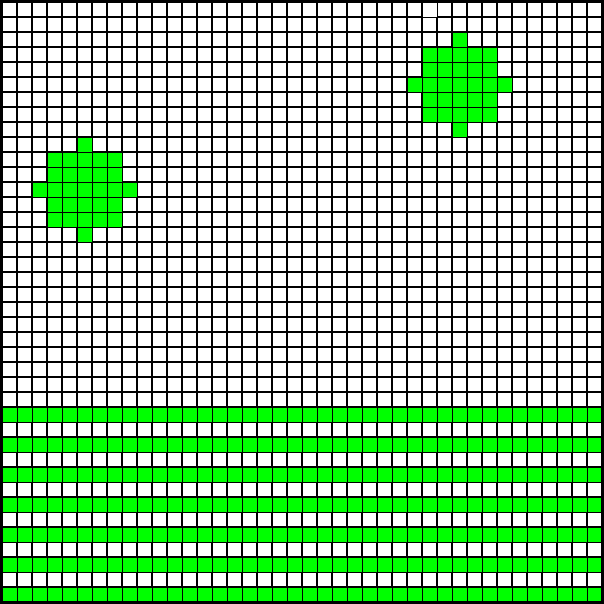}};
        %To scale the dimensions of brace in case if image is scaled in future
        \begin{scope}[x={(image.south east)},y={(image.north west)}]
          \node[] at (-0.2,0.15) {\textbf{$Z_2$}};
          \draw[thick, orange, decorate, decoration={brace, amplitude=5pt}] (-0.04,0)--(-0.04,0.33);
          %drawing the arrows
          \node (A) at (-0.2, 0.42) {$Z_1$};
          \node (B) at (0.8, 0.85) {};
          \draw[orange, very thick, ->] (B) edge (A);
          \node (C) at (-0.2, 0.7) {$Z_0$};
          \node (D) at (0.2, 0.71) {};
          \draw[orange, very thick, ->] (D) edge (C);
        \end{scope}
      \end{tikzpicture}
    }
    \subfloat[]
    {
      \begin{tikzpicture}
        \node[anchor=south west,inner sep=0] (image) at (0,0)
        {\includegraphics[width=25mm]{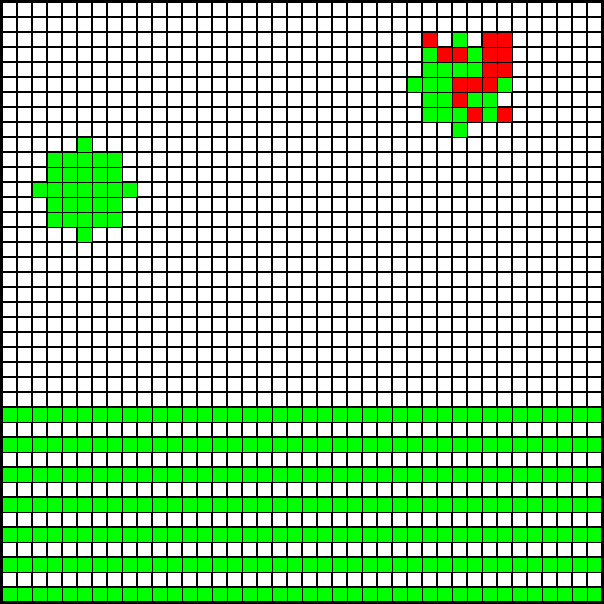}};
      \end{tikzpicture}
    }\\
    \vspace*{-0.3cm}
    \hspace*{0.7cm}
    \subfloat[]
    {
      \begin{tikzpicture}
        \node[anchor=south west,inner sep=0] (image) at (0,0)
        {\includegraphics[width=25mm]{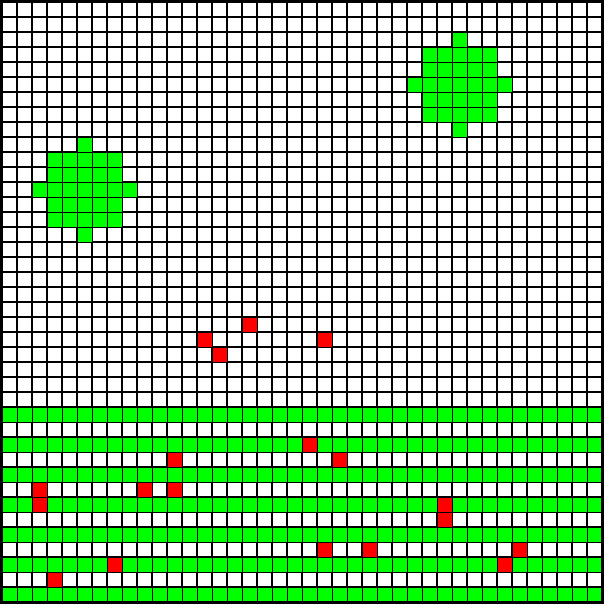}};
      \end{tikzpicture}
    }
  \subfloat[]
  {
    \begin{tikzpicture}
      \node[anchor=south west,inner sep=0] (image) at (0,0)
      {\includegraphics[width=25mm]{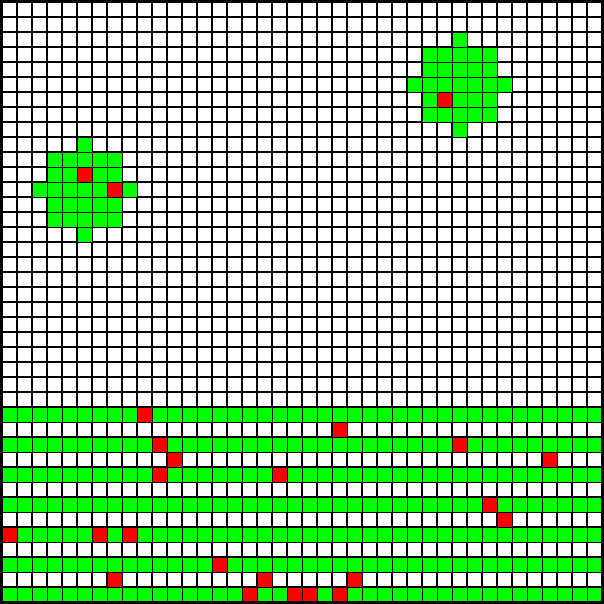}};
    \end{tikzpicture}
  }
    \caption{\small{\textbf{(a)} The toy environment distribution used to demonstrate the complementarity between the two fitnesses. \textbf{(b)} Results of a run in which only adaptivity ($f_1$) is optimized. \textbf{(c)} results of a single run where only polyvalence ($f_0$) is maximized. \textbf{(d)} Results when both $f_0$ and $f_1$ are jointly maximized. See section \ref{sec_ablation} for details.}}
    \label{ablation_fig}
\end{figure}
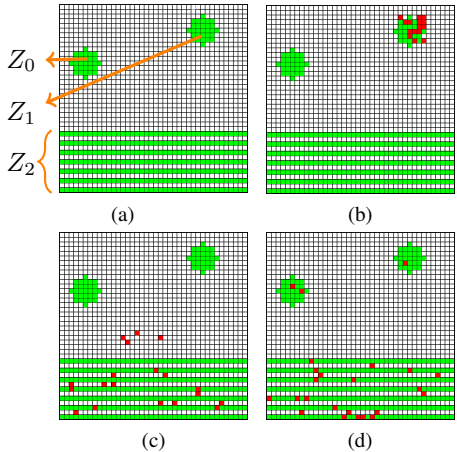

The choice of source and task distributions can be guided by metric functions defined between (PO)MDPs. However, those metrics often have shortcomings. For example, distance functions based on generative models \cite{ammar2014automated} are dependent on exploration, and graph-based approaches \cite{song2016measuring} are limited to finite state spaces. As a result, as in most of the current Reinforcement Learning literature, we let intuition guide our choice of source and task targets. In this experiment, we chose \texttt{basketball-v2} as the source task for transfer, and consider \texttt{shelf-place-v2}, \texttt{pick-place-v2} and \texttt{bin-picking-v2} as the target tasks\footnote{We did not use the ML10 benchmark from metaworld, since except for \texttt{place-shelf-v2}, the tasks that compose that benchmark present very little challenge for QD methods, and we preferred to select more difficult tasks.}. The latter three environments had proven to be difficult to solve in a reasonable number of generations using QD optimization, mainly because of the high reward returned for brittle policies that exploit the particularities and bugs in the physics simulator, making the occurrence of proper grasping behaviors rare. The aims of \texttt{pick-place-v2} and \texttt{shelf-place-v2} are, as the names indicate, to pick an object and place it respectively at a given $3d$ location in space and on a shelf. The goal of \texttt{bin-picking-v2} is to pick up items from a bin and place them in another one.

The goal of the source task is to throw a ball into a hoop, and successful policies learn to grasp the ball before moving the end-effector and releasing it towards the hoop. As the three difficult discussed environments require similar behaviors, it seems likely that some of the knowledge gained from the basketball environment will be transferable to them. 

The results of our experiments are reported in table \ref{table_ctg}. For each of the target tasks, as indicated in the \textit{prior type}, three experiments are considered: 1) QD optimization without priors 2) QD optimization with priors that consist of $\mu$ random solutions to $\mu$ environments from $\mathcal{T}_{source}$ and 3) QD optimization using priors learned by FAERY. In each case, the results are computed based on $135$ samples from the corresponding task distribution.

Those results clearly indicate that the use of priors significantly facilitates optimization. In particular, no solutions were found for \texttt{shelf-place-v2} in a reasonable number of generations ($G_{QD}=2000$) without using priors. While this does indeed show that \texttt{basketball-v2} is a suitable source task, we see that there is a difference in experiments that were initialized with solvers from $\mathcal{T}$ and those that were initialized with the FAERY priors. The latter seem to constantly outperform the former. However, the margin between the two range from small on \texttt{shelf-place-v2} to significant on \texttt{pick-place-v2} and \texttt{bin-picking-v2}. Note that as in previous tables and figures, the mean number of generations does not take into account environments that a method is not able to solve and for which $G_{QD}$ generations are nonetheless wasted. It therefore understates the reduction in learning time that results from utilizing the learned priors. While more exhaustive experiments should be carried in the future, those results seem to hint at the potential of our method for generalization across different tasks.
\subsection{Ablation Study.}
\label{sec_ablation}
We demonstrate the complementarity of the two fitnesses $f_0, f_1$ on a toy environment distribution whose simplicity allows us to provide further insights via visualization. Each environment from this distribution is a $40\times40$ grid, where a single cell $(g_1,g_2)$ is designated as the goal, which an agent is expected to correctly guess in a single action \footnote{Each environment from this distribution can be seen as a multi-armed bandit problem with a horizon of $1$ and binary rewards that are null everywhere except for a single action.}. The distribution of those environments is illustrated in figure \ref{ablation_fig}(a). The green areas, noted $Z_0, Z_1, Z_2$ in that figure, indicate the cells that can be chosen as goal. In other words, an environment is instantiated by taking an empty $40\times40$ grid together with a single goal cell $(g_1,g_2)$ chosen from the green areas. For simplicity, each agent is parameterized by the single action $(i,j)$ it can take. This action is also used as the behavior descriptor. The experiment then consists in running FAERY with $\lambda=\mu=25$ for $70$ generations, initializing  all agents randomly on the first row of the grid, and using half the cells from of each $Z_i$ area as the training set. Mutation is realized by random increments to the agents in the left, right, up and down directions. We ran the experiment $15$ times.

As shown in figure \ref{ablation_fig}(b), when optimizing only for adaptivity ($f_1$ from eq. \ref{eq_updates}), the meta-population gets trapped in the $Z_1$ area without exploring the other two. This means that while the resulting population is able to quickly adapt to environments sampled from $Z_1$, none of its individuals are able to do so for $Z_0$ and $Z_2$. Optimizing only for polyvalence ($f_0$ from eq. \ref{eq_updates}), however, moves the meta-population towards the parts where the distribution of tasks has more mass, abandoning other areas. This is illustrated in figure \ref{ablation_fig}(c), where the population is clearly drawn to $Z_2$ while ignoring the two other areas. Finally, as shown in figure \ref{ablation_fig}, optimizing both $f_0, f_1$ jointly allows the population to be well-distributed among all three task distribution modes.

Note that figures \ref{ablation_fig} (b,c,d) illustrate the most likely outcome that we observed during the different runs, but the stochasticity of the mutation and selection operators (in breaking ties) can lead to slightly different results. While in all experiments, optimizing only $f_1$ results in a situation where the meta-population remains in $Z_1$ and optimizing only $f_0$ leaves at least one of $Z_1$ or $Z_2$ uncovered, in $13.3\%$ of experiments, optimizing the two fitnesses jointly results in a meta-population that has individuals close to but not inside $Z_0$.

\section{Conclusion}
\label{sec_concl}

We presented an algorithm that leverages previous QD experience to enable Few-Shot Quality Diversity optimization on previously unseen tasks. It does not require dense rewards nor gradients and is agnostic to the underlying QD optimization method, as well as to the models used to represent policies. Furthermore, it is easy to implement and scale. The empirical validations that were presented in single-task generalization settings indicate that our approach not only considerably reduces the number of generations required for QD optimization on a new problem, but also hint that it could be a promising direction for multi-task generalization.

\bibliographystyle{IEEEtran}
\bibliography{IEEEabrv,example}

\end{small}
\end{document}